# Toward a new instances of NELL

Maísa C. Duarte and Pierre Maret

Univ. Lyon, UJM-Saint-Etienne, CNRS, Laboratoire Hubert Curien UMR 5516
F-42023 Saint Etienne, France

**Abstract:** We are developing the method to start new instances of NELL in various languages and develop then NELL's multilingualism. We base our method on our experience on NELL Portuguese and NELL French. This reports explain our method and develops some research perspectives.

## 1. Introduction

NELL (Never-Ending Language Learner) [1] is a computer system that runs 24 hours per day, 7 days per week. It was started up on January, 12th, 2010 and should be running forever, reading the web and gathering more and more facts to grow and populate its own knowledge base.

In short, we can describe the system as follows: NELL's initial knowledge base (KB) is an ontology defining hundreds of categories (e.g., athlete, sports, sportsTeam, fruit, product, country, city, emotion, etc.) and relations (e.g., athletePlaysForTeam (athlete, sportsTeam), cityLocatedInCountry (city, country)) and a set of 10 to 15 examples (instances) for each one of the categories (e.g. athlete (Kobe Bryant), sportsTeam(LA Lakers), etc.) and for each one of the relations (e.g. athletePlaysForTeam (Kobe Bryant, LA Lakers), cityLocatedInCountry (New York, USA), etc.). Publications that explain in more detail about NELL and that support the current project can be found in [1], [4], [5] and [6]. The publication [7] is the last publication about all the system and currently components.

The standard process of NELL is depicted in Figure 1 in a simple and generic view. As shown in this figure, the input is the all-pairs-data and the ontology/knowledge base (KB), the output is the ontology/knowledge base. Let's explain better each part of the process.

The present document describes what is necessary to setting up a new NELL instance in a different language. In resume, it is necessary an ontology and an input, both are described in details in the Sections 2 ,3 and 4. The process to create a new NELL instance were published for the first time for the Portuguese NELL version ([3, 2]), and for the second time for the French version. This document is based on these publications and experiences.

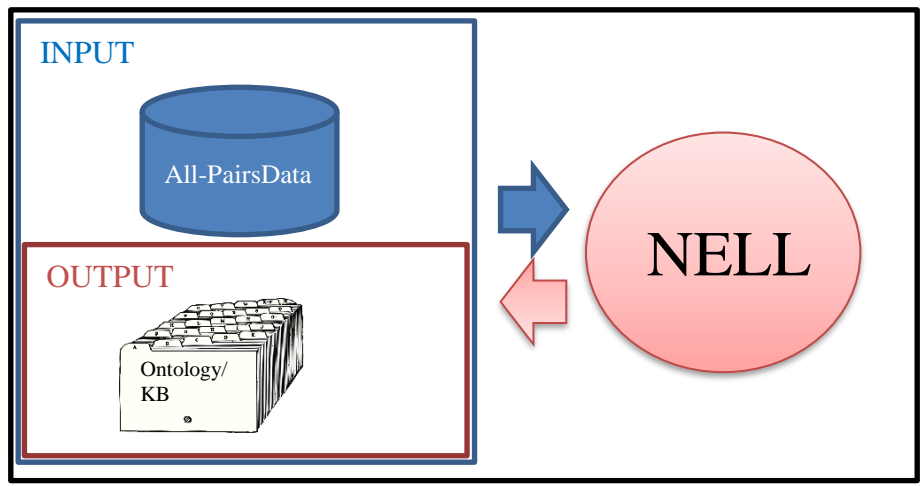

**Fig. 1. The NELL global process**

## 2. NELL's input

The input of NELL is the web. NELL reads and learn from webpages. NELL learns all time, "forever". NELL's key is to read and understand better each day. We can imagine a human reading a book on an unknown subject, and as much the human reads the book, as much knowledge he is able to extract. NELL does a similar task: it reads the web a lot of times (call "iterations"), which requires a lot of time and resources. From the text, NELL pre-processes a source base called **All-Pairs-Data.**

## 3. A preprocessed source base: All-Pairs-Data

An All-Paris-Data is made using a big corpus and stores all occurrences and co-occurrences between **Named Entities** (**NE**) and **Textual Patterns** (**TP**) in two views: **Categories** and **Relations**.

Categories is the learning of a unary relation. For example the Category City can have the unary relation: City(New York). Relations is the learning of a binary relation. For example the Relation LocatedIn(New York,USA). For Categories the system extracts just one instance for predicate (New York), while for relations it extracts one pair of instances for predicate (New York, USA).

An All-Pairs-Data is created for Categories and Relations or just for Categories. For Categories the all-pairs will consist of all occurrences between a TP and a NE. For Relations all occurrences between a pair of NE's and a TP.

It is shown in the Tables 1 and 2 one simple example for an All-Pairs-Data of Category (Table 1) and another of Relation (Table 2). In these tables we have the number of occurrences between NE and TP for categories, and between a pair of NEs and a TP for relations. When NELL learns, NELL makes the math to discover and count the co-occurrence.

*Table 1 - Example of All-Pairs-Data for Categories*

| Categories All-Pairs-Data | | |
|---|---|---|
| NE | TP | Occurrence |
| Saint Étienne | is a city | 20 |
| São Paulo | is a city | 30 |
| New York | is a city | 50 |
| Saint Étienne | is a beautiful city | 30 |
| Saint Étienne | city such as | 30 |
| São Paulo | city such as | 60 |
| New York | city such as | 100 |

*Table 2 - Example of All-Pairs-Data for Relations*

| Relations All-Pairs-Data | | | |
|---|---|---|---|
| NE | NE | TP | Occurrence |
| Saint Étienne | France | is located in | 20 |
| São Paulo | Brazil | is located in | 20 |
| New York | USA | is located in | 40 |
| Saint Étienne | France | is the most beautiful city in | 20 |
| Paris | France | is the capital of | 30 |
| Brasília | Brazil | is the capital of | 30 |

The NE and TP extractions are guided by a *part-of-speech / Tagging* process. Any other approaches can be applied for extracted the NE and TP. The only important thing is that the NE and TP identified are not modified. In other words, it's important to keep the *strings* as it was found on the original text. The point is that it extracts exactly what is written in order to keep the process consistent (for more information about the keys of NELL access: http://www.cs.cmu.edu/)

Currently, some of the approaches applied on NELL to create an All-Pairs for Categories is: after the NE is found, two TP are extracted, one on the left of the NE and the other on the right of the NE. For example: "Located in USA, New York is a very famous city". For the NE "New York" it will be extracted one TP on the left: "Located in USA,". And another on the right: "is a very famous city". Five (5) grams are used to extract the TP in English.

For other languages (basically all languages, except English for now) combinations between 2 or 3 and 5 grams are saved. For example: "is a city", "is a city located", "is a city located near". For English a filter is used to find the better TP for a NE.

In French and Portuguese, for Relations, the TP is the sentence between the NE pair. For categories, the number of grams is between 3 and 5. For English the number of grams is also until 5, but there are different filters to find the best TP. These filters have not been developed for other languages yet.

## 4. NELL's Ontology

In order to start a new NELL instance, we need to give names of some Categories and Relations, examples of them and some configuration parameters to keep the structure of the ontology. For this we've created spreadsheets which are then interpreted by script programs. The execution of the script creates the NELL initial Knowledge Base (KB) and this initial KB is used in the learning process in order to learn more (*input* and *output* of the NELL process).

### 4.1 Ontology Mapping

In order to create a new instance of NELL in a new language, we can create an ontology from scratch, or we can map the English one. Ontology Mapping is the process when the English spreadsheets are used to make a mapping for another language. We say "mapping" because it's not just a translation because categories names and relations names may have to be adapted for the new language. For example: State (original category in English, translated "Etat" in French) may have a different translation depending on the context.

All the columns of the spreadsheet have to be mapped. The most important points are 1) give the names of categories and relations in the new language and 2) give seeds for each predicate. Some examples of the spreadsheet columns for categories are: category name, seeds of instances (e.g.: for the category city the seeds can be: Paris, Saint-Etienne, Lyon, etc.), human format (e.g.: "X is a city", where X is the learned city), mutex exceptions that define what categories are *not* mutual exclusive (e.g.: the category politician has the mutex exceptions: professor, astronaut, celebrity and so on), etc.

For the relation spreadsheet we have the almost the same columns of categories, but there is some additional information about the NE part. For example, the column "number of values" must have the value 1 or N to describe the number of connections that the NE on the right of the relation can have with the NE on the left. In the same way there is the column "number of inverse values" for the NE on the left.

Some examples for the spreadsheet of relation are: relation name, seeds of instances (e.g.: for the relation ceoOf, some seeds can be: "Sundar Pichai, Google", "Tim Cook, Apple", etc.), human format (e.g.: "X is the ceo of Y", where X refers the instance of the category person and the Y refers the instance of a category company), mutex exceptions (e.g.: for buildingLocatedInCity the mutex exception is the relation aquariumInCity), and the columns "number of values" and "number of inverse values" (for the relation ceoOf the values are 1 and 1, because one company has just one ceo, and one ceo works just for one company.

### 4.2 Human Supervision

After the new ontology is created, it can be used in the NELL process. Then it is possible to start the human supervision step.

The human supervision on the knowledge base is used to make NELL learning better and faster. The system learns every day and it needs to be supervised and corrected in order to learn rejecting wrong interpretations of text. The supervision task has some parameters (time/number of supervision) that can be changed with the evolution of the learning process.

## 5. Challenges

We are developing the method to start new instances of NELL in various languages and develop then NELL's multilingualism. We base our method on our experience on NELL Portuguese and NELL French.

A next step will be to work on the trans-lingual aspects of NELL. How to export the knowledge from English version for the other ones? The idea is that we can improve the knowledge base of all NELL versions if we share and exchange the knowledge from different languages. Such "bridges" between the knowledge bases would help the new ones to grow up faster and then to learn and improve faster. Reversely, text in other languages may indirectly enrich the English NELL which is not be able to find certain knowledge.

Another important perspective is to provide an easy way to map the NELL ontology with another knowledge base, for example DBPedia. BDPedia is based on Semantic Web techniques and makes intensive use of RDF (Semantic Web knowledge representation model). We work on applications that are able to generate RDF from NELL, or reversely to feed NELL from RDF knowledge bases. Other representation models can be used to represent knowledge. Our approach will facilitate the current and future interactions and integration efforts between NELL and other knowledge bases.